\documentclass[conference]{IEEEtran}
\IEEEoverridecommandlockouts
\usepackage{cite}
\usepackage{amsmath,amssymb,amsfonts}
\usepackage{algorithmic}
\usepackage{graphicx}
\usepackage{textcomp}
\usepackage{xcolor}
\usepackage{color,soul}

\def\BibTeX{{\rm B\kern-.05em{\sc i\kern-.025em b}\kern-.08em
    T\kern-.1667em\lower.7ex\hbox{E}\kern-.125emX}}
\begin{document}

\title{Harnessing Deep Learning and Satellite Imagery for Post-Buyout Land Cover Mapping}

%\author{\IEEEauthorblockN{Anonymous Authors}}
\author{%
  \IEEEauthorblockN{%
    Hakan T. Otal\IEEEauthorrefmark{1},
    Elyse Zavar\IEEEauthorrefmark{2},
    Sherri B. Binder\IEEEauthorrefmark{3}, 
    Alex Greer\IEEEauthorrefmark{4}, and
    M. Abdullah Canbaz\IEEEauthorrefmark{1}
  }%
\IEEEauthorblockA{ \IEEEauthorrefmark{1} Department of Information Sciences and Technology College of Emergency Preparedness, Homeland Security, and Cybersecurity \\ \textit{University at Albany, SUNY, Albany, NY}\\ \textit{ hotal, mcanbaz [at] albany [dot] edu;}}%

\IEEEauthorblockA{ \IEEEauthorrefmark{2} Department of Emergency Management and Disaster Science \\ \textit{University of North Texas}\\ \textit{ elyse [dot] zavar [at] unt [dot] edu;}}%

\IEEEauthorblockA{ \IEEEauthorrefmark{3} \textit{BrokoppBinder Research \& Consulting, LLC, Allentown, PA}\\ \textit{ sbinder [at] brokoppbinder [dot] com}}%

\IEEEauthorblockA{ \IEEEauthorrefmark{4} Department of Emergency Preparedness and Homeland Security\\ College of Emergency Preparedness, Homeland Security, and Cybersecurity \\ \textit{University at Albany, SUNY, Albany, NY}\\ \textit{ agreer [at] albany [dot] edu;}}%

 \vspace{-4.6mm}
}

\maketitle

\begin{abstract}
Environmental disasters such as floods, hurricanes, and wildfires have increasingly threatened communities worldwide, prompting various mitigation strategies. Among these, property buyouts have emerged as a prominent approach to reducing vulnerability to future disasters. This strategy involves governments purchasing at-risk properties from willing sellers and converting the land into open space, ostensibly reducing future disaster risk and impact. However, the aftermath of these buyouts, particularly concerning land-use patterns and community impacts, remains under-explored. This research aims to fill this gap by employing innovative techniques like satellite imagery analysis and deep learning to study these patterns. To achieve this goal, we employed FEMA's Hazard Mitigation Grant Program (HMGP) buyout dataset, encompassing over 41,004 addresses of these buyout properties from 1989 to 2017. Leveraging Google's Maps Static API, we gathered 40,053 satellite images corresponding to these buyout lands. Subsequently, we implemented five cutting-edge machine learning models to evaluate their performance in classifying land cover types. Notably, this task involved multi-class classification, and our model achieved an outstanding ROC-AUC score of 98.86\% 
\end{abstract}


\vspace{-2mm}

\section{Introduction}
Increasing awareness of the risks and impacts of climate change and other related hazards is prompting communities to consider tools for mitigating risks to people and property. Among these options, property acquisitions, including home buyout programs, are often considered as a means of permanently eliminating hazard risk \cite{greer2022hazard}. Home buyout programs provide federal funding for state or local government agencies to purchase properties that are identified as being at risk for significant or repeated hazard losses. Decisions about the design and implementation of these programs are made locally, including decisions regarding which properties to include or exclude. In most cases, the programs are considered voluntary, meaning each individual property owner has the option to participate in or opt out of the program \cite{greer2017historical}. Participating property owners sell their properties to the overseeing government agency, after which any structures on the property are demolished, and the government agency is required to maintain the purchased properties as open space in perpetuity. Though these programs result in the relocation of residents out of hazardous areas, significant challenges arise with the management and maintenance of the open space that is created when properties are depopulated. Though these properties have the potential to be maintained as open spaces with high ecological utility, such as forested areas or rain gardens, they most often remain vacant lots with minimal ecological or social value \cite{zavar2016land}. However, to date, limited data on post-buyout land uses has made comprehensive assessments of land use decisions and utility challenging, thus hindering efforts to improve land use decision-making and outcomes.

Understanding the complex interplay between economic, social, and environmental factors influencing land use decisions after buyouts is crucial. These factors grapple with the immediate economic benefits versus long-term fiscal impacts \cite{curran} \cite{binder}, the psychological and community-centric consequences of displacement\cite{devries}, and the effects of policy decisions and natural recovery processes on environmental outcomes\cite{zavar2020, liu}. Analyzing these factors objectively requires a comprehensive understanding of post-buyout land use.

%While existing research provides valuable insights into buyouts as a tool for disaster recovery and hazard mitigation \cite{munoz_unequal_2016,fraser_creating_2006}, concerns regarding their effectiveness remain under-addressed in the mainstream \cite{smith_planning_2011, burkett_reaching_2017}. Addressing these concerns requires a robust data-driven approach that leverages AI-enhanced decision support and image classification tasks. By analyzing extensive post-buyout land use data, this approach can systematically assess the interplay between various factors, leading to a more informed understanding of buyouts' effectiveness.

From an emerging technological standpoint, there are two crucial concepts directly related to the Buyout land problem. First, the ever-increasing availability of high-resolution satellite imagery presents both exciting opportunities and significant challenges in the fields of image classification and land use segmentation \cite{lesiv2018characterizing,malarvizhi2016use}. A major challenge lies in the vastness and complexity of data, demanding efficient and scalable processing algorithms \cite{papadopoulos2021hard}. Additionally, the heterogeneity of landscapes and the presence of mixed pixels necessitate robust approaches to handling diverse spectral and spatial characteristics \cite{mishra2019performance}.

Second, in the realm of artificial intelligence, research efforts are actively exploring deep learning techniques, particularly convolutional neural networks (CNNs)\cite{ouchra2021satellite, kadhim2020convolutional} and semantic segmentation models\cite{neupane2021deep}, to achieve more accurate and automated land use classification. One crucial research topic involves developing methods for robust feature extraction that can effectively capture the subtle variations within complex land cover types\cite{tsai2002derivative}. Another area of investigation focuses on incorporating multi-temporal data to enhance classification accuracy and track changes in land use over time \cite{kumar2020multi}. 

To this end, the expansion of buyout research through a data-driven approach holds significant potential for advancing our comprehension of these programs and their effectiveness, particularly in the context of climate change, global warming, and risk management and mitigation. Leveraging AI-enhanced decision support and image classification tasks to scrutinize post-buyout land use data allows for valuable insights that can inform policy decisions, leading to improved outcomes for communities affected by disasters and hazards. This paper presents a novel approach to advance buyout research by leveraging data-driven techniques and AI-powered decision support. We utilize a comprehensive dataset of over 41,000 buyout properties acquired through the FEMA Hazard Mitigation Grant Program (HMGP), spanning a period of nearly three decades. Building upon this data, we employ Google's Maps Static API to gather high-resolution satellite imagery for 40,053 buyout locations. These data resources are then integrated with five cutting-edge machine learning models to assess their effectiveness in classifying land cover types. This innovative application of data science and AI has the potential to significantly enhance our understanding of buyout programs and their impact on land use patterns.

%\newpage
\vspace{-1mm}
\section{Methodology}

\vspace{-2mm}
\subsection{Datasets}

In this study, two primary datasets were utilized. The first, essential for understanding post-buyout land-use patterns, is the FEMA HMGP Dataset\cite{openfema}. This dataset comprises properties acquired through buyouts as part of the Hazard Mitigation Grant Program and provided the necessary addresses for analysis, encompassing over 41,004 addresses for the buyout properties.

Secondly, the UCMerced Land Use Dataset\cite{ucmerced} was used, a publicly accessible collection curated by the University of California. This 256x256 pixel dataset includes 2,100 images for 17 classes and was crucial in training the machine learning model. Due to the richness and diversity of this dataset, a strong model was created, capable of accurately classifying different types of land use patterns.

\subsection{Preprocessing}

\begin{figure}[!ht]
\centerline{\includegraphics[width=\linewidth]{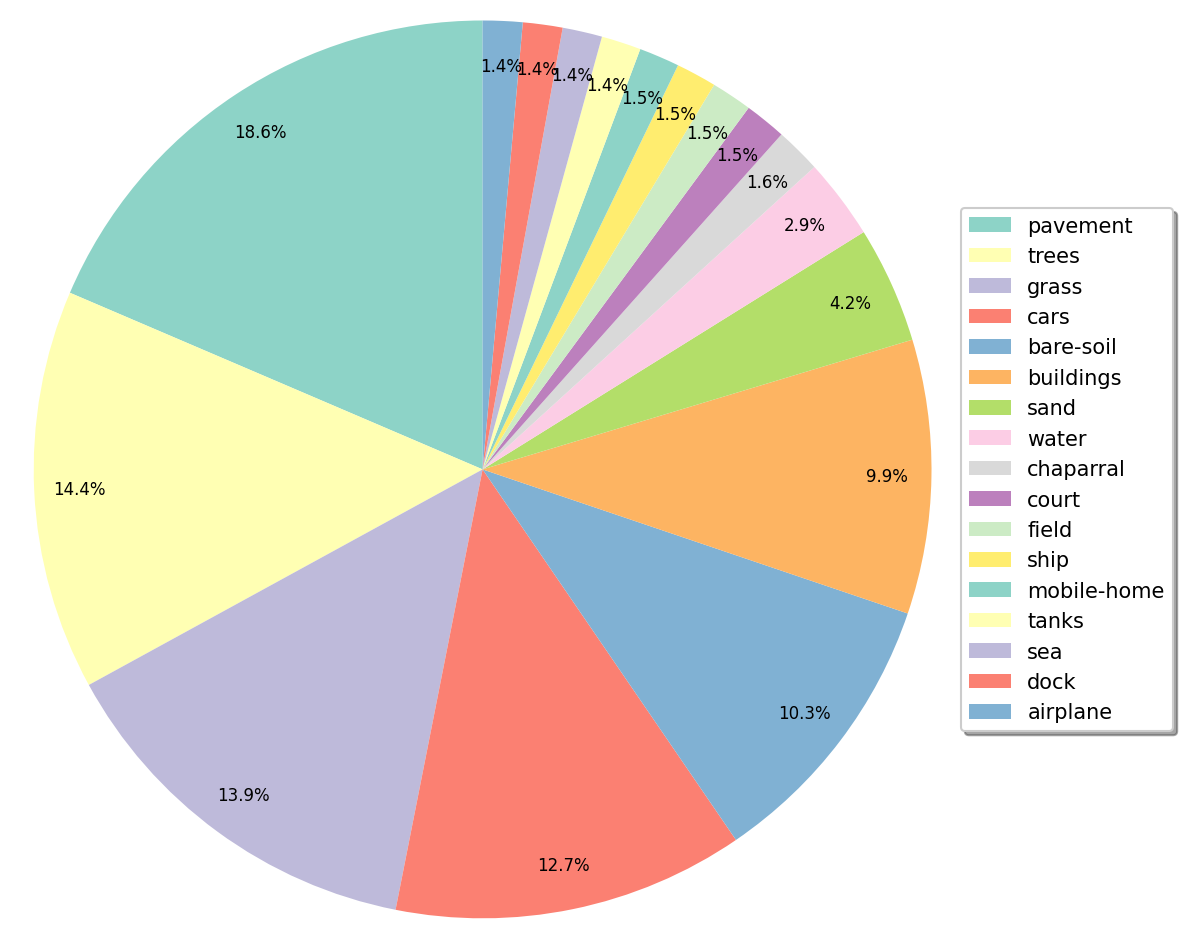}}
\caption{Land use class distribution of UC Merced Dataset (training data)}
\label{fig:trainpiechart}
\end{figure}

There are 17 land use classes in the UC Merced Land Use Dataset, and it has been preprocessed to fit a training format. The land use class distribution of the training dataset is shown in Figure \ref{fig:trainpiechart}. Image augmentation techniques such as flips and rotations are used to increase the dataset's variability in order to improve the models' performance.

Please note that all images used in training phase were in 256x256 pixels format for all models except Inception-v3 model. Since Inception-v3's architecture needs training images to be at least 299x299 pixels, all images in the dataset were upscaled to 299x299 pixels only for training of Inception-v3 model.

Utilizing the addresses from the FEMA HMGP dataset, satellite images were harvested via the Google Maps API\cite{mapsapi}. It is important to note that of the 41,004 addresses in the FEMA HMGP dataset, only 40,053 satellite images were retrieved from the Google Maps API query due to address errors. This discrepancy is attributed to incorrect or unidentified addresses in the FEMA-provided dataset. The images, captured in 640x640 pixels, served as the primary visual data for the analysis. Preprocessing steps were also conducted on these images, resizing them to match the format of the training dataset.
Please keep in mind that in this work, the models mentioned above were trained with the UC Merced Land Use Dataset and tested with the FEMA HMGP dataset.

\begin{figure*}[!htb]
    \centering
    \begin{minipage}{.5\textwidth}
        \centering
        \includegraphics[width=.9\linewidth]{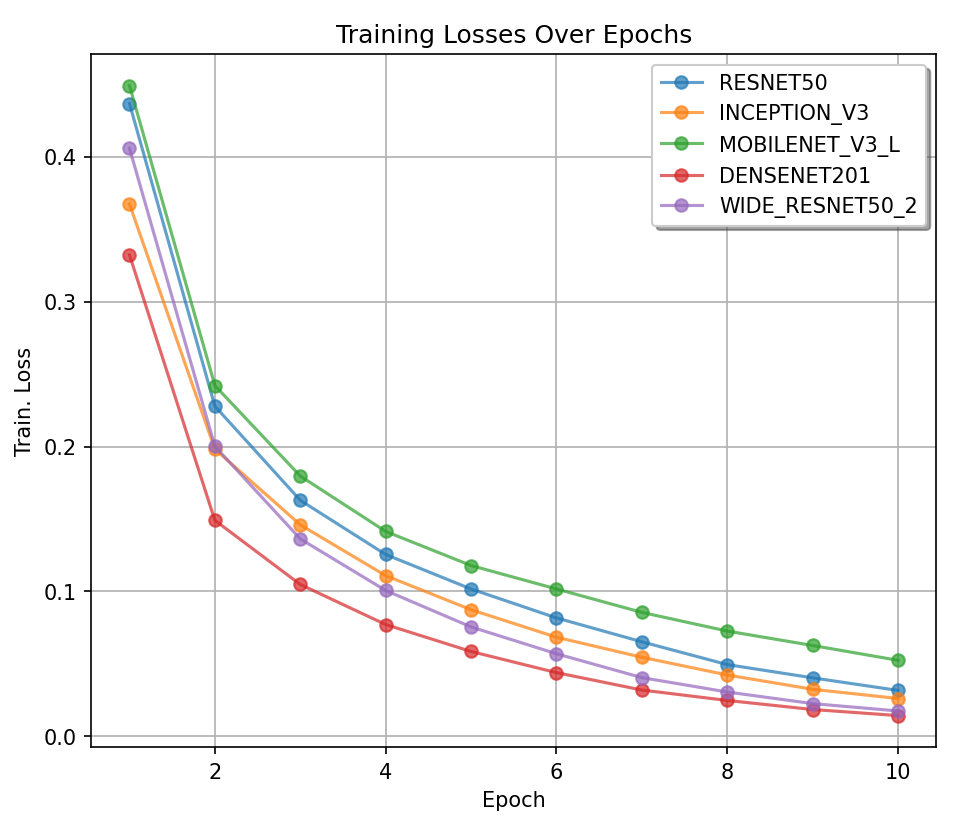}
        \caption{Training losses of different computer vision models over epochs}
        \label{fig:trainloss}
    \end{minipage}%
    \begin{minipage}{0.5\textwidth}
        \centering
        \includegraphics[width=.9\linewidth]{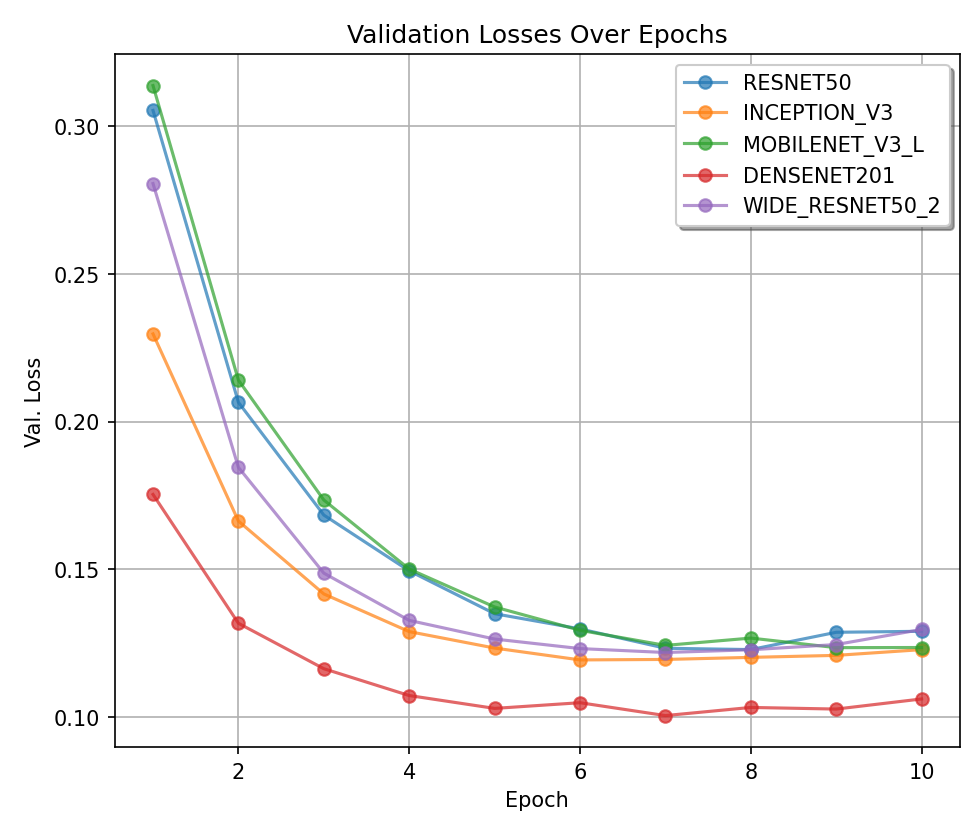}
        \caption{Validation losses of different computer vision models over epochs}
        \label{fig:valloss}
    \end{minipage}
 %   \vspace{-4mm}
\end{figure*}
%\vspace{2mm}

\subsection{Image Classification Models}

The approach to analyzing land-use patterns involved training deep learning models using the preprocessed UC Merced Land Use Dataset. Experiments were conducted to identify the most efficient neural network architecture for the objective, utilizing a range of vision models: Resnet-50\cite{he_deep_2015}, Inception-V3\cite{szegedy_rethinking_2015}, MobileNet-V3\cite{howard_searching_2019}, Densenet-201\cite{huang_densely_2018}, and WideResnet-50 \cite{zagoruyko_wide_2017}.

\textbf{Resnet-50} model is known for its "skip connections" or "shortcut connections" that skip one or more layers. In Resnet-50, these connections help in avoiding the problem of vanishing gradients by allowing this alternate pathway for the gradient to flow through. It has 50 layers deep and is more efficient in terms of computation and memory compared to deeper models.

Part of the Inception family, \textbf{Inception-V3} model is known for its efficiency and lower computational cost. It uses modules called "Inception modules" which allow it to look at the same image with different receptive fields (sizes of filters), enabling it to capture spatial hierarchies in data more effectively. Inception-V3 introduced several improvements over its predecessors, including factorizing convolutions and better utilization of the computing resources inside the network.

\textbf{MobileNet-V3} model is designed specifically for mobile and resource-constrained environments. It is highly efficient in terms of both size and performance. MobileNet-V3 uses lightweight depthwise separable convolutions and incorporates architecture search techniques and a squeeze-and-excitation optimization, making it a very compact and efficient model for mobile devices.

\textbf{DenseNet} architectures are unique because they connect each layer to every other layer in a feed-forward fashion. For each layer, the feature maps of all preceding layers are used as inputs, and its own feature maps are used as inputs into all subsequent layers. This leads to substantial reductions in the number of parameters and improves efficiency.

Wide Residual Networks are a variation of Resnet where the width (number of channels) of the network is increased while decreasing its depth. \textbf{WideResnet-50}, in particular, is a version with 50 layers. This change in architecture helps in improving the model's accuracy while keeping the computational budget almost the same.

For those unfamiliar with technical jargon, the training process involved monitoring training and validation losses and adjusting parameters to optimize the model's performance. For training Python programming language, Pytorch and Pandas libraries are used. 

In Figures \ref{fig:trainloss} and \ref{fig:valloss}, the training and validation losses of the tested models are shown. The training loss plot depicts a steady decrease over the ten epochs for all models, indicating successful model learning. The rate of decrease can be analyzed for potential optimization opportunities, but it's not our concern for this work.
The validation loss plot for our models follows a similar trend as the training loss but remains slightly higher. There are minor fluctuations starting at epoch 6, which suggests overfitting. Also, this indicates that the models can generalize well to unseen data. Based on the training and validation loss figures, \textbf{\textit{DenseNet201}} appears to be the best model among those tested as it demonstrates the lowest overall loss and best generalization capabilities.

\subsection{Best Model Selection}

In the study, the multi-classification task is central to understanding post-buyout land-use patterns. This task involves categorizing satellite imagery into predefined classes of land use. The complexity of this task arises from the nuanced differences between land use types, which can often be subtle and overlap in visual characteristics. For instance, a single land image might simultaneously encompass trees, cars, and pavement. To accurately gauge the model's performance, a range of metrics, including accuracy, F1-score, and ROC-AUC, was employed. These metrics are crucial as they provide a comprehensive understanding of the model's strengths and weaknesses.

The most straightforward metric, \textit{accuracy}, is expressed as the percentage of accurate forecasts among all predictions made. However, accuracy alone can be deceptive in a multi-class context, especially if the dataset is unbalanced. To better understand the model's capacity to accurately identify each class, considering any biases or imbalances in the dataset, accuracy was supplemented with other measures.

Another statistical metric called the \textit{F1-score} is used in binary classification to assess how well recall and precision are balanced. It is the harmonic mean of recall, which measures the percentage of real positives among all positive predictions, and precision, which measures the percentage of genuine positives among all positive forecasts.

A graphical plot called the \textit{Receiver Operating Characteristic (ROC)} curve shows how well a binary classifier system can diagnose problems when its discriminating threshold is changed.
The model's level of separability is shown by the Area Under the Curve (AUC). It indicates the degree to which the model can discriminate between classes. Better model performance is indicated by higher AUC values. A flawless classifier is represented by a ROC-AUC score of 1, whereas a fully random classifier is represented by a score of 0.5. This score is very helpful for assessing classifiers on datasets that are unbalanced.This multi-faceted approach to metrics ensures a well-rounded evaluation of the model's performance in the multi-classification task.

In Table \ref{tab:metrics}, we present a comparative analysis of various models including Resnet-50, Inception-V3, Mobilenet-V3, Densenet-201, and WideResnet-50, focusing on key metrics such as accuracy, precision, recall, F1 score, and ROC-AUC score. While Mobilenet-V3 leads in terms of accuracy and Resnet-50 excels in precision, WideResnet-50 shows superior performance in recall. Notably, Densenet-201 emerges as the most effective model overall, achieving the highest scores in both the F1 and ROC-AUC metrics, which are crucial for a comprehensive evaluation of model performance.

\begin{table}[!t]
\renewcommand{\arraystretch}{1.2}
\caption{Metrics of the Tested Models}
\label{tab:metrics}
\centering
\begin{tabular}{|c||c|c|c|c|c|}
\hline
Model & Accuracy & Precision & Recall & F1 & ROC-AUC\\
\hline\hline
Resnet-50 & 51.67 & 90.23 & 86.13 & 88.13 & 98.69 \\
Inception-V3 & 49.52 & 85.53 & 90.79 & 88.08 & 98.73 \\
Mobilenet-V3 & 51.90 & 88.72 & 87.47 & 88.09 & 98.67 \\
Densenet-201 & 51.67 & 88.46 & 90.16 & 89.30 & 98.86 \\
WideResnet-50 & 51.19 & 85.23 & 93.36 & 89.11 & 98.76 \\
\hline
\end{tabular}
\end{table}

Please note that the selection of an appropriate model depends on specific performance priorities. Models such as Resnet-50 and Densenet-201 are suitable when emphasizing precision and accuracy. For an emphasis on recall, WideResnet-50 exhibits the highest recall. When seeking a balance between precision and recall, the F1 score serves as the primary metric, and Densenet-201 exhibits the highest F1 score. For assessing the model's discriminatory capacity in binary classification, ROC-AUC is pertinent, and Densenet-201 exhibits the highest ROC-AUC.

The choice of DenseNet201 as the architecture for goal for classifying the buyout land covers was driven by several considerations. DenseNet201 is a convolutional neural network known for its dense connectivity pattern, where each layer is connected to every other layer in a feed-forward fashion. This architecture is particularly beneficial for our task for several reasons. Firstly, DenseNet201 exhibits improved information flow and feature propagation, making it highly efficient in learning distinguishing features from land use images. This is crucial for our multi-classification task, where the differentiation of subtle land-use features is key. Secondly, DenseNet201 is also known for its parameter efficiency, reducing the risk of overfitting while handling a large number of input images. Its architecture allows it to require fewer parameters than other networks of similar depth, making it a more computationally efficient choice. Finally, DenseNet201 has demonstrated superior performance in various image classification tasks both in previous studies and in our tests as depicted in Table \ref{tab:metrics}, indicating its potential effectiveness in accurately classifying land use from satellite imagery. These factors collectively informed our decision to leverage DenseNet201 for this intricate task of land use classification.

\begin{figure}[!ht]
\centerline{\includegraphics[width=\linewidth]{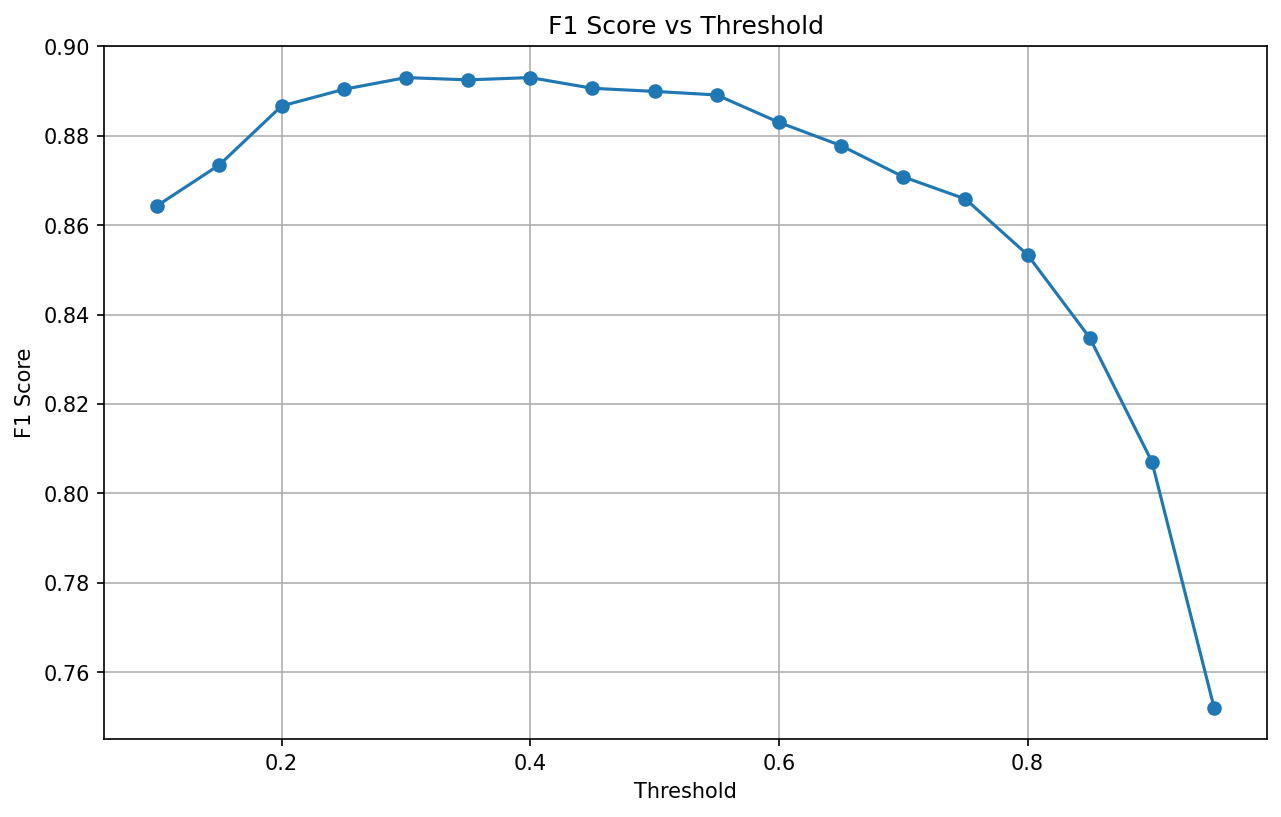}}
\vspace{-2mm}
\caption{DenseNet201's F1-scores over various confidence thresholds}
\label{fig:threshold}
\vspace{-2mm}
\end{figure}

\begin{figure*} [!htb]
\centerline{\includegraphics[width=0.98\linewidth]{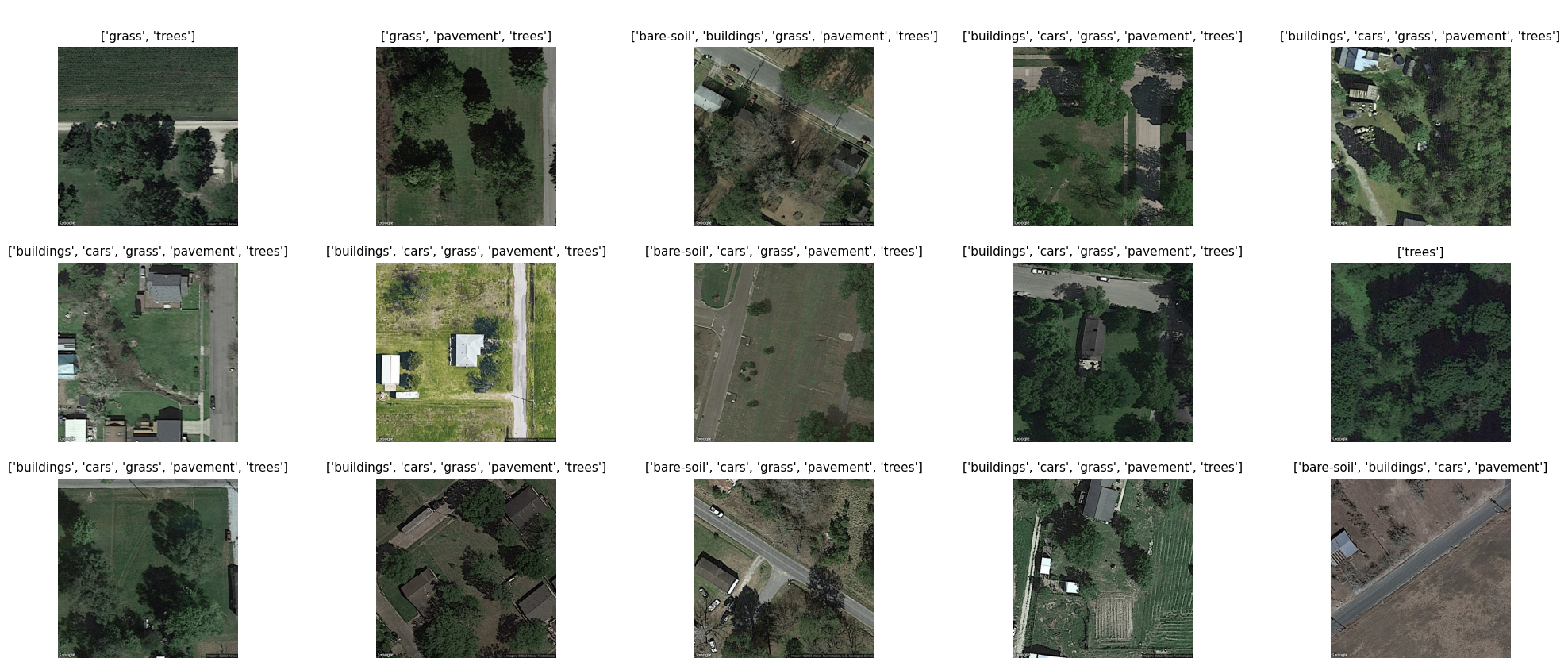}}
\caption{The sample of model's land use class predictions of collected satellite images}
\label{fig:satellitepredictions}
\end{figure*}

\subsection{Best Confidence Threshold}

Another aspect of the methodology was establishing the best confidence threshold for the model's predictions. Specifically, the \textit{best confidence threshold} refers to the cutoff point for accepting or rejecting predictions made by a machine learning model, balancing the trade-off between accuracy and completeness. This involved an iterative process of balancing the need for accuracy with the practicality of prediction confidence. In Figure \ref{fig:threshold}, F1-scores for different threshold values were tested and $0.4$ was identified as the best. The chosen threshold ensured that the model's predictions on images were reliable and could be used confidently in further analysis.

\subsection{Hardware Requirements}
The training and deployment of the deep learning models demanded significant computational resources. By harnessing NVIDIA cloud GPU services' robust processing capabilities, it was possible to effectively continue the model training process and they proved to be abundantly sufficient for the successful completion of this project.

\section{Experimental Results}

In this section, we provide experimental results of the land cover classification through satellite image analysis. Our primary focus is to methodically categorize distinct land cover types, such as vegetated areas, impervious surfaces, water bodies, and transportation infrastructure.

Figure \ref{fig:satellitepredictions} illustrates a small sample of the model's land use class predictions of collected satellite images. Our model has identified several distinct land cover categories, including vegetated zones (trees), impermeable structures (buildings, pavement), aquatic environments (including sandy regions and sea areas), and elements of transportation infrastructure (cars). Each image in the study was labeled to reflect the dominant land cover types it displayed.

Upon scrutinizing the data, a notable predominance of impermeable surfaces and transportation-related infrastructure was observed in land parcels acquired through FEMA's buyout program dataset. It is important to acknowledge that these parcels have previously experienced significant flooding events or other disasters. Presently, these areas are characterized by the presence of non-porous structures and expansive parking facilities, which collectively signify a suboptimal use of urban land.

\begin{figure}[!b]
\centerline{\includegraphics[width=\linewidth]{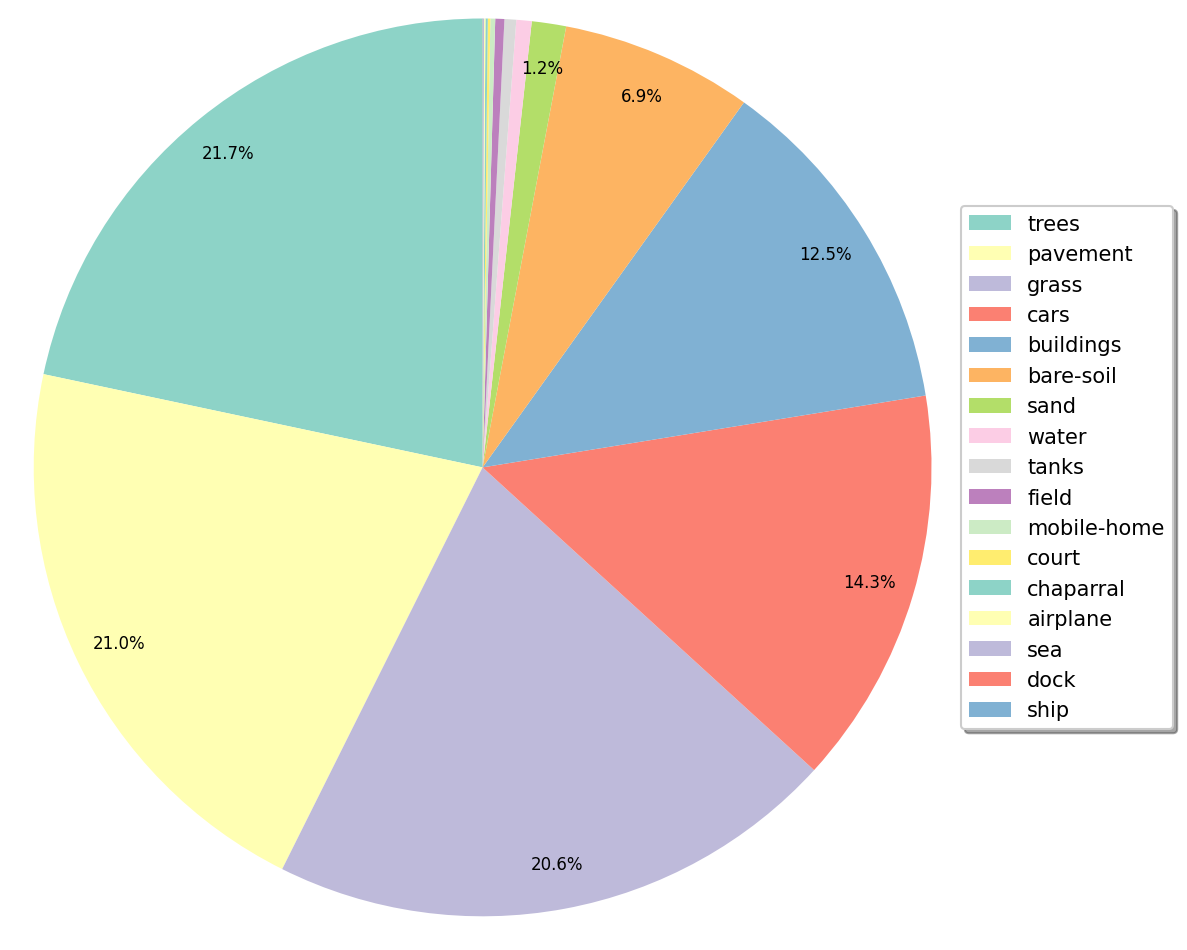}}
\caption{Distribution of DenseNet-detected land use classes of satellite images}
\label{fig:satellitepiechart}
\end{figure}

In figure \ref{fig:satellitepiechart}, we present a pie chart illustrating the distribution of various land cover types detected within land areas purchased through FEMA's buyout program, presumably following a significant environmental incident. The land cover classes include natural features such as trees, grass, and water bodies, as well as anthropogenic elements like buildings, pavement, and vehicles.

From the image classification and machine learning point of view, this chart can be interpreted as a visual representation of the current state of land utilization within these buyout areas. Also, please note that this is a multi-class classification problem, as shown in Figure \ref{fig:satellitepredictions}, as there are multiple classes detected in some of the images. A significant proportion of the land is occupied by impervious surfaces such as buildings (14.3\%) and pavement (21.0\%), and vehicles (cars at 12.5\%). The chart serves as an empirical basis for discussing the need to reassess and redirect land use strategies in disaster-affected buyout lands towards more sustainable and community-focused purposes. Hence, we provide an additional explanation of these findings from the domain experts in the next subsection.

\begin{figure}[!b]
\centerline{\includegraphics[width=\linewidth]{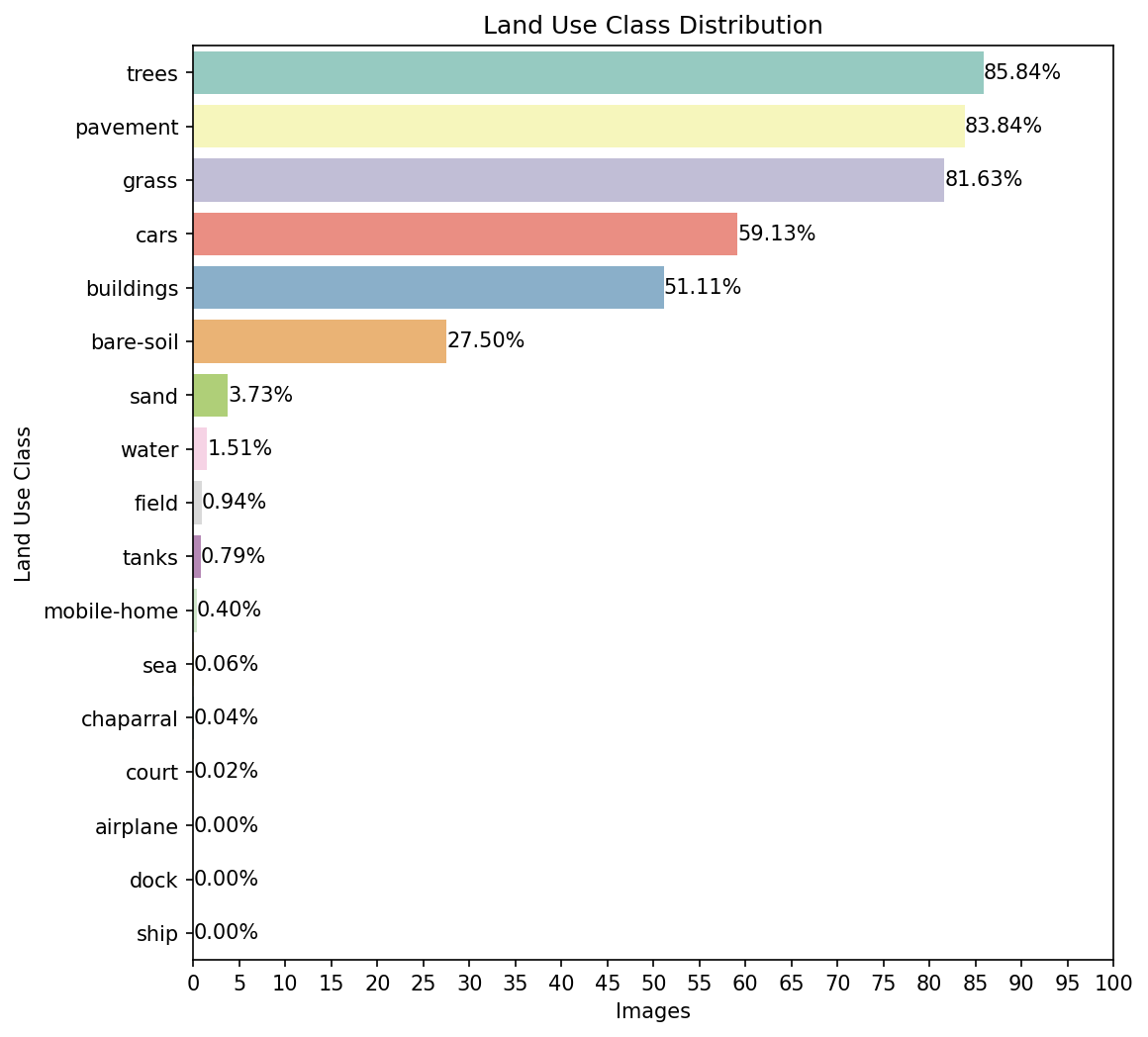}}
\caption{Distribution of land use classes on satellite images}
\label{fig:satellitebargraph}
\end{figure}

Similarly, in figure \ref{fig:satellitebargraph}, we show the bar graph of the distribution of identified land cover classes in areas acquired through buyout programs after a major incident. The horizontal axis represents the percentage of images in which each land use class was detected, while the vertical axis lists the different land use classes.

From the image classification perspective, the graph highlights that the majority of the land cover is dominated by natural vegetation, with trees (85.84\%), grass (81.63\%), and bare soil (27.50\%) being most prevalent. This is consistent with the recommended use of these lands as green spaces. However, there is a significant presence of impermeable surfaces such as pavement (83.84\%) and buildings (51.11\%), as well as a notable quantity of cars (59.13\%). This suggests that a substantial portion of the land is covered by non-permeable, artificial structures, rather than being converted into green spaces or recreational areas as would be ideal for regions at risk of environmental hazards.

% \begin{figure}[!htb]
% \centerline{\includegraphics[width=\linewidth]{figures/satellite_images_confidences.png}}
% \caption{The model's mean confidence for each land use class}
% \label{fig:satelliteconfidences}
% \end{figure}

\subsection{Interpreting the Results}

The deep learning models trained using the UC Merced Land Use Dataset accurately identified post-buyout land uses from Google Earth satellite imagery. Most of the images classified contain trees (21.7\%), pavement (21.0\%), grass (20.6\%), cars (14.3\%), and buildings (12.5\%), with a total of 90.1\% as depicted in figure \ref{fig:satellitepiechart}. This is consistent with previous work on post-buyout land use which found that most lots remain mowed yet vacant a decade after acquisition \cite{dascher2023biophilia}\cite{zavar2016land}.  However, to identify these uses previously, extensive fieldwork was required. With the deep learning models, post-buyout land uses can be identified without resource-intensive site visits and offer accurate results. The high frequency of grass in the classified satellite images likely reflects the continued occurrence of mowed, vacant lots as the dominant post-buyout land use. Most communities conducting buyouts seek to retain existing trees, or institute tree plantings on buyout lots thus contributing to the high frequency of trees observed in the satellite image classification. The high frequency of pavement reported in the image classification is due in part to the satellite image containing roads adjacent to buyout properties. Communities also commonly leave driveway cuts and foundations on vacant buyout lots which are likely contributing to some of the pavement detected in the satellite images. Research has also identified cars commonly parked on post-buyout vacant lots, especially by adjacent neighbors \cite{zavar2023expression}. The presence of buildings in the classified images are likely due to buyout lots occurring in neighborhoods with homes remaining, a land use pattern referred to as checkerboarding \cite{binder2020home}. While there is a preference for acquiring contiguous lots, as the U.S. buyout programs represented in the FEMA HMGP dataset were voluntary, homeowners could decline to participate in the program, leaving housing interspaced with buyout lots. However, one constraint of the Google Earth satellite images is that the image is not restricted to the buyout parcel boundaries and, therefore, contains elements from adjacent properties or streets. 
Importantly, the model was able to distinguish between vacant lot land cover types, including grass, trees, and bare soil, as depicted in \ref{fig:satellitebargraph}, which illustrates a small sample of the model’s land use class predictions of collected satellite images. These various land covers, all common to vacant lots, offer differing degrees of hazard risk reduction. For example, while trees can reduce the effects of flash flooding and abate poor air quality and extreme heat \cite{asah2012involving}, these ecosystem services are reduced at locations dominated by grass or bare soil. It can be difficult to identify all land cover types during fieldwork as many communities utilize physical barriers like cement roadblocks used to reduce access to buyout lots. Thus, the aerial view from the satellite images and model identification can offer a more nuanced and detailed account of land cover.  
 
\section{Conclusion}

In this paper, we introduce a novel methodology that significantly enhances the analysis of land use in post-buyout properties, integrating machine learning research and the development of task-specific deep learning vision models. Traditional approaches to this challenge have relied on labor-intensive and costly in-person site assessments. In contrast, our study leverages state-of-the-art deep learning algorithms, fine-tuned on the UC Merced Land Use Dataset, and employs high-resolution Google Earth satellite imagery to accurately classify the land uses of properties acquired through buyout programs. This technological approach is particularly pertinent as climate change intensifies and buyouts become more critical in our environmental disaster mitigation strategies. Comprehending how these lands are utilized post-buyout is essential for evaluating the impact of such interventions on community resilience.

Looking ahead, future research should aim to refine this approach through several advancements. Firstly, the scope of land use classifications within our machine learning model should be broadened. The current training dataset, while precise, does not encompass all possible post-buyout land uses, such as agricultural areas, wetlands, playgrounds, detention basins, and athletic fields. These categories have been identified as common post-buyout uses in prior research and thus represent a crucial expansion for the model's training data to improve its accuracy and applicability.

Secondly, there is a need for enhanced techniques to isolate the satellite imagery analysis to the specific plots acquired through buyouts, thereby ensuring that the data is not confounded by adjacent non-buyout areas like roads or neighboring properties. This could involve the development of more sophisticated image segmentation models that can discern and focus exclusively on the boundaries of buyout properties.

Additionally, the integration of temporal data analysis could also be valuable. By tracking changes in land use over time with sequential satellite images, we can gain insights into the evolution of post-buyout land use and its long-term implications for community resilience.

Finally, the integration of these advancements necessitates not only technical improvements in model architecture and training data but also a multidisciplinary approach that incorporates insights from urban planning, environmental science, and policy analysis to inform model development. This comprehensive perspective is crucial to ensure that the deep learning models we develop are not only technically proficient but also contextually aware and aligned with the overarching goals of sustainable land use and disaster resilience.  

\section{Acknowledgements}

This material is based upon work supported by the Google Cloud Research Credits program with the award GCP19980904.

%\newpage
\bibliographystyle{IEEEtran}
\bibliography{paper}

\end{document}